\documentclass[11pt, a4paper]{article}
\usepackage{acl2015}
\usepackage{times}
\usepackage{url}
\usepackage{latexsym}
\usepackage{graphicx} 
\usepackage{amsmath}
\usepackage{amssymb}
\usepackage{subfigure}
\usepackage{multirow}
\usepackage{rotating}
\usepackage[small]{caption}
\usepackage{float} 

\def\noDA{\textsc{Static}}
\def\batch{\textsc{Batch}}
\def\online{\textsc{Online}}
\def\figref#1{Figure~\ref{fig:#1}}
\def\figlabel#1{\label{fig:#1}\label{p:#1}}

\def\tabref#1{Table~\ref{tab:#1}}
\def\tablabel#1{\label{tab:#1}\label{p:#1}}

\def\seclabel#1{\label{sec:#1}\label{p:#1}}
\def\eqref#1{Eq.~\ref{eqn:#1}}

\def\uprm#1{\mbox{$^{\hbox{\scriptsize #1}}$}}

\def\panelfactor{0.37}
\DeclareMathOperator{\tf}{tf}
\DeclareMathOperator{\freq}{freq}
\DeclareMathOperator{\bigram}{bigram}

\newcommand{\enotesoff}{\long\gdef\enote##1##2{}}

\enotesoff

\long\def\eat#1{\ignorespaces}

\title{Online Updating of Word Representations for Part-of-Speech Tagging}
\author{}
\author{
Wenpeng Yin \\
  LMU Munich \\
  {\tt wenpeng@cis.lmu.de} \\\And
Tobias Schnabel \\
  Cornell University \\
  {\tt tbs49@cornell.edu} \\\And
  Hinrich Sch\"{u}tze \\
  LMU Munich \\
  {\tt inquiries@cislmu.org} \\}
\date{}

\begin{document}
\maketitle
\begin{abstract}
We propose \emph{online
  unsupervised domain
  adaptation (DA)}, which is performed
\emph{incrementally} as data comes in and is applicable when
batch DA is not possible.
In a part-of-speech (POS) tagging evaluation,
we
find that online unsupervised DA
performs as well as
batch DA.
\end{abstract}


\enote{tm}{

 - (Introduction) I would like to see an example where you
really do
 not have target domain data.
 Crawling doesn't seem ideal. Something like chat in social
networks,
 where you really do not know
 what people will talk about?
 
 - (Introduction) Is the last paragraph not revealing that
your are the
 FLORS authors?
 
 -(Experimental Setup): How much TD data are you using
 
 - (Experimental Setup): Sounds a little weird to me: ``one
each for its
 suffix, its shape and its left and right distributional
neighbors.''   

}

\enote{tm}{

train\_flors.py

}

\enote{ts}{

process tobias' comments

}

\section{Introduction}
\emph{Unsupervised domain adaptation} is a scenario that practitioners often face when having to build robust NLP systems. They have labeled data in the source domain, but wish to improve performance in the target domain by making use of \emph{unlabeled data} alone.
Most work on unsupervised domain adaptation in NLP uses \emph{batch learning}: It  assumes that a large corpus of unlabeled data of the target domain is available before testing. 
However, batch learning is not possible in many real-world scenarios where incoming data from a new target domain must be processed immediately. More importantly, in many real-world scenarios the data does not come with neat domain labels and it may  not
be immediately obvious that an input stream is suddenly
delivering data from a new domain.

Consider
an NLP system
that analyzes emails at an enterprise. There is a constant
stream of incoming emails and it changes over time --
without any clear indication that the models in use should
be adapted to the new data distribution. Because the system
needs to work in real-time, it is also desirable to do any
adaptation of the system \emph{online}, without the
need of stopping the system, changing it and restarting it
as is done in batch mode.

In this paper, we propose \emph{online unsupervised domain adaptation}
as an extension to traditional unsupervised DA. In online unsupervised DA, domain adaptation is performed
incrementally as data comes in. Specifically, we adopt a
form of \emph{representation learning}. In our experiments, the
incremental updating will be performed for representations
of words. Each time a word is encountered in the stream of data at test time, its representation is updated.

To the best of our knowledge, the work reported here is the
first study of online unsupervised DA. More specifically, we
evaluate online unsupervised DA for the task of POS
tagging. We compare POS tagging results for three distinct
approaches: static (the baseline), batch learning and
online unsupervised DA. Our results show that online
unsupervised DA is comparable in performance to batch
learning while requiring no retraining or prior data in the
target domain.


\begin{table*}[!t]
\centering
\setlength{\tabcolsep}{3pt}
{\footnotesize
\begin{tabular}{l|l@{\hspace{0.1cm}}r|l@{\hspace{0.2cm}}r|l@{\hspace{0.2cm}}r|l@{\hspace{0.2cm}}r|l@{\hspace{0.2cm}}r|l@{\hspace{0.2cm}}r}
      &   \multicolumn{2}{c|}{newsgroups} & \multicolumn{2}{c|}{reviews}&  \multicolumn{2}{c|}{weblogs}&  \multicolumn{2}{c|}{answers}&  \multicolumn{2}{c|}{emails} & \multicolumn{2}{c}{wsj}\\
      &   \multicolumn{1}{c}{ALL}&\multicolumn{1}{c|}{OOV}&\multicolumn{1}{c}{ALL}&\multicolumn{1}{c|}{OOV}&\multicolumn{1}{c}{ALL}&\multicolumn{1}{c|}{OOV}&\multicolumn{1}{c}{ALL}&\multicolumn{1}{c|}{OOV}&\multicolumn{1}{c}{ALL}&\multicolumn{1}{c|}{OOV}&\multicolumn{1}{c}{ALL}&\multicolumn{1}{c}{OOV}\\\hline
TnT & 88.66 & 54.73 & 90.40 & 56.75 & 93.33 & 74.17 & 88.55 & 48.32 & 88.14 & 58.09 &95.75 & 88.30\\
Stanford & 89.11 & 56.02 & 91.43 & 58.66 & 94.15 & 77.13 & 88.92 & 49.30 & 88.68 & 58.42&\textbf{96.83} & 90.25\\
SVMTool & 89.14 & 53.82 & 91.30 & 54.20 & 94.21 & 76.44 &88.96 & 47.25 & 88.64 & 56.37 &96.63 & 87.96\\
C\&P & 89.51 & 57.23 & 91.58 & 59.67 & 94.41 & 78.46 &89.08 & 48.46 & 88.74 & 58.62 &96.78 & 88.65\\
S\&S &90.86 &66.42&92.95&75.29&94.71&83.64&90.30 & 62.16 & 89.44 & 62.61&96.59&90.37\\
S\&S (reimpl.) & 90.68 & 65.52 & 93.00 & 75.50 & 94.64 & 82.91 & 90.18 & 61.98 & 89.53 & 62.46 & 96.60 & 89.70\\\hline
\batch{} & \textbf{90.87}&\textbf{71.18}&\textbf{93.07}&\textbf{79.03}&\textbf{94.86}&\textbf{86.53}&\textbf{90.70}&\textbf{65.29}&89.84&65.44&96.63&\textbf{91.86}\\
\online{} & 90.85&71.00&\textbf{93.07}&\textbf{79.03}&\textbf{94.86}&\textbf{86.53}&90.68&65.16&\textbf{89.85}&\textbf{65.48}&96.62&91.69
\end{tabular}}

\caption{\batch{} and
\online{}
accuracies are
  comparable and state-of-the-art. Best number in each
  column is bold.}\tablabel{comp}
\end{table*}

\section{Experimental setup}\label{experiment}
\seclabel{setup}

\textbf{Tagger.}
We reimplemented the   FLORS
tagger \cite{schnabel13flors}, a fast and simple tagger that
performs well in DA. It treats POS tagging as a window-based (as opposed to
sequence classification), multi\-label classification problem. FLORS is ideally suited for
online unsupervised DA because its representation of words includes
distributional vectors -- these vectors can be easily updated
in both batch learning and online unsupervised DA.
More specifically, a word's representation in FLORS consists of four
feature vectors: one each for its suffix, its shape and its
left and right distributional neighbors.
Suffix and shape features are standard features used in the
literature; our use of them is exactly as described
by \newcite{schnabel13flors}. 

\emph{Distributional features.} 
The $i\uprm{th}$ entry $x_i$ of the left distributional
vector of $w$ 
is the weighted number of times  the \emph{indicator word} 
$c_i$ occurs immediately to the left
 of $w$:\\
\mbox{\hspace{15mm}}$x_i = \tf\left(\freq\left(\bigram(c_i, w)\right)\right)$\\
where $c_i$ is the word with frequency rank $i$ in the
corpus,
$\freq\left(\bigram(c_i, w)\right)$ is the number of
occurrences of the bigram
``$c_i$ $w$''  and
we weight non-zero frequencies logarithmically: $\tf(x) = 1 +
\log(x)$.
The right distributional vector is defined
analogously. 
We restrict the set of indicator words to the $n=500$ most frequent words.
To avoid zero vectors,
we add an entry $x_{n+1}$ to each vector that counts  omitted contexts:\\
\mbox{\hspace{8mm}}$x_{501} = \tf(\sum_{j: j>n} \freq\left(\bigram(c_j,
w)\right))$

\enote{wy}{
FLORS defines a left distributional vector for $w$ to count
the times of top 500 frequent words in the corpus occurring
immediately to the left of $w$, and each non-zero raw entry
is reweighted by $1+log(\cdot)$. The right distributional
vector is defined analogously. To avoid zero vectors, an
extra entry $x_{501}$ is added to count omitted context.
}

\enote{hs}{
When computing distributional vectors, case distinctions are
ignored; thus, ``Bush'' and ``bush'' have the same
distributional vectors. Case distinctions are
preserved for suffix and shape features.}

Let $f(w)$ be the
concatentation of the two distributional  and suffix
and shape vectors of word $w$.
Then FLORS represents token $v_i$ as follows:\\
\mbox{\hspace{5mm}}$f(v_{i-2}) \oplus f(v_{i-1}) \oplus f(v_{i})
\oplus f(v_{i+1}) \oplus f(v_{i+2})$\\
where $\oplus$ is vector concatenation. FLORS then tags token $v_i$ based on this
representation.

FLORS assumes that the association between distributional
features and labels does not change fundamentally when going
from source to target. This is in contrast to other work,
notably 
\newcite{blitzer06structural},  that carefully selects
``stable'' distributional features and discards ``unstable''
distributional features. The hypothesis underlying FLORS is
that basic distributional POS properties are relatively stable across domains -- in contrast to
semantic and other more complex tasks. The high performance
of FLORS 
\cite{schnabel13flors} suggests this hypothesis is true.


\textbf{Data.}
\emph{Test set.} We evaluate on the development sets of six different TDs:
five SANCL \cite{petrov2012overview} domains -- newsgroups, weblogs,
reviews, answers, emails -- and sections 22-23 of WSJ for in-domain testing.

We use two \emph{training sets}
of different sizes. In condition \emph{l:big} (big labeled data set), we train FLORS on sections 2-21 of Wall Street Journal (WSJ). Condition \emph{l:small} uses 10\% of l:big.

\emph{Data for word representations.}  We also vary the size
of the datasets that are used to compute the word
representations before the FLORS model is trained on the
training set.
In condition \emph{u:big}, we compute distributional
vectors on the joint corpus of all labeled and unlabeled
text of source and target domains (except for the test sets).
We also include 100,000
WSJ sentences from 1988 and 500,000 sentences from Gigaword
\cite{parker2009english}. In condition \emph{u:0}, only
labeled training data is used.

\textbf{Methods.}
We implemented the following modification compared to the
setup in \cite{schnabel13flors}: distributional vectors are kept
in memory as count vectors. 
This allows us to increase the counts during online tagging.

We run experiments with three versions of FLORS: \noDA{},
\batch{} and \online{}. All three methods compute word
representations on ``data for word representations''
(described above) before the model is trained on one of the
two ``training
sets'' (described above).

\noDA{}.
Word representations are not changed during testing.


\batch{}. Before testing,
we update  count vectors by $\freq\left(\bigram(c_i, w)\right)$ += $\freq^*\left(\bigram(c_i, w)\right)$,
where $\freq^*(\cdot)$ denotes the number of occurrences of
the bigram ``$c_i$ $w$'' in the entire test set.

\begin{table*}[!t]
\centering
\setlength{\tabcolsep}{3pt}
{\footnotesize
\begin{tabular}{lll|llll|llll}
&      &      &       \multicolumn{4}{c|}{u:0} & \multicolumn{4}{c}{u:big}\\
&      &      &   ALL & KN & SHFT & OOV & ALL &KN & SHFT & OOV\\\hline\hline
\multirow{6}{*}{\begin{sideways}{newsgroups} \end{sideways}} 
&\multirow{3}{*}{\begin{sideways}{l:small} \end{sideways}}
    & \noDA{} &87.02 &90.87&71.12&57.16&89.02&91.48&81.53&58.30\\
&    & \online{} &{87.99}&{90.87}&{76.10}&{65.64}&\textbf{89.84}&\textbf{92.38}&{82.58}&\textbf{67.09}\\
&    & \batch{} &\textbf{88.28} &\textbf{91.08}&\textbf{77.01}&\textbf{66.37}&{89.82}&{92.37}&\textbf{82.65}&{67.03}\\\cline{2-11}
&\multirow{3}{*}{\begin{sideways}{l:big} \end{sideways}}
    & \noDA{} &89.69 &93.00&82.65&57.82&89.93&92.41&84.94&58.97\\
  &  & \online{} &{90.51} &\textbf{93.13}&82.51&{67.57}&{90.85}&\textbf{93.04}&{84.94}&{71.00}\\
   & & \batch{} &\textbf{90.69} &{93.12}&\textbf{83.24}&\textbf{69.43}&\textbf{90.87}&{93.03}&\textbf{85.20}&\textbf{71.18}\\\hline
\multirow{6}{*}{\begin{sideways}{reviews} \end{sideways}} 
&\multirow{3}{*}{\begin{sideways}{l:small} \end{sideways}}
    & \noDA{}  
&89.08&91.96&66.55&65.90&91.45&92.47&80.11&70.81\\
&    & \online{}
&{89.67}&{92.14}&\textbf{70.14}&{69.67}&\textbf{92.11}&\textbf{93.62}&{81.46}&\textbf{78.42}\\
&    & \batch{} 
&\textbf{89.79}&\textbf{92.23}&{69.86}&\textbf{71.27}&{92.10}&{93.60}&\textbf{81.51}&\textbf{78.42}\\\cline{2-11}
&\multirow{3}{*}{\begin{sideways}{l:big} \end{sideways}}
    & \noDA{}
&91.96&93.94&82.30&67.97&92.42&93.53&84.65&69.97\\
&    & \online{} 
&{92.33}&{94.03}&\textbf{83.59}&{72.50}&\textbf{93.07}&\textbf{94.36}&\textbf{85.71}&\textbf{79.03}\\
&    & \batch{}
&\textbf{92.42}&\textbf{94.09}&{83.53}&\textbf{73.35}&\textbf{93.07}&\textbf{94.36}&\textbf{85.71}&\textbf{79.03}\\\hline
\multirow{6}{*}{\begin{sideways}{weblogs} \end{sideways}} 
&\multirow{3}{*}{\begin{sideways}{l:small} \end{sideways}}
    & \noDA{}  
&91.58&94.29&79.95&72.74&93.42&94.77&89.80&77.42\\
&    & \online{}
&{92.51}&{94.52}&{81.76}&{80.46}&\textbf{94.21}&{95.40}&\textbf{91.08}&\textbf{84.03}\\
&    & \batch{} 
&\textbf{92.68}&\textbf{94.60}&\textbf{82.34}&\textbf{81.20}&{94.20}&\textbf{95.42}&{91.03}&{83.87}\\\cline{2-11}
&\multirow{3}{*}{\begin{sideways}{l:big} \end{sideways}}
    & \noDA{}
&93.45&95.64&\textbf{90.15}&72.68&94.09&95.54&91.90&76.94\\
&    & \online{} 
&{94.18}&{95.82}&89.80&{80.35}&\textbf{94.86}&{95.81}&\textbf{92.60}&\textbf{86.53}\\
&    & \batch{}
&\textbf{94.34}&\textbf{95.85}&90.03&\textbf{81.84}&\textbf{94.86}&\textbf{95.82}&\textbf{92.60}&\textbf{86.53}\\\hline
\multirow{6}{*}{\begin{sideways}{answers} \end{sideways}} 
&\multirow{3}{*}{\begin{sideways}{l:small} \end{sideways}}
    & \noDA{}  
&86.93&90.89 &66.51& 53.43& 88.98& 91.09 &77.63& 57.36\\
&    & \online{}
&{87.48}& \textbf{91.18}&{68.07}& {56.47}& \textbf{89.71}& {92.42} &{78.11}& \textbf{64.21}\\
&    & \batch{} 
&\textbf{87.56}& {91.11}& \textbf{68.25}& \textbf{58.44}& \textbf{89.71}& \textbf{92.43}& \textbf{78.23}& {64.09}\\\cline{2-11}
&\multirow{3}{*}{\begin{sideways}{l:big} \end{sideways}}
    & \noDA{}
&89.54& 92.76& 78.65& 56.22& 90.06& 92.18& 80.70& 58.25\\
&    & \online{} 
&{89.98}& {92.97}& \textbf{79.07} &{59.77}& {90.68}& {93.21}& {81.48}& {65.16}\\
&    & \batch{}
&\textbf{90.14}& \textbf{93.10} &{79.01}& \textbf{60.72}& \textbf{90.70} &\textbf{93.22} &\textbf{81.54}& \textbf{65.29}\\\hline

\multirow{6}{*}{\begin{sideways}{emails} \end{sideways}} 
&\multirow{3}{*}{\begin{sideways}{l:small} \end{sideways}}
    & \noDA{}  
&85.43& 90.85& 57.85& 51.65 &87.76& 90.35& 70.86& 56.76\\
&    & \online{}
&{86.30} &{91.26}& {60.56}& {55.83} &{88.45}& {92.31}& {71.67}& {61.57}\\
&    & \batch{} 
&\textbf{86.42}& \textbf{91.31}& \textbf{61.03}& \textbf{56.32} &\textbf{88.46} &\textbf{92.32}& \textbf{71.71}& \textbf{61.65}\\\cline{2-11}
&\multirow{3}{*}{\begin{sideways}{l:big} \end{sideways}}
    & \noDA{}
&88.31& 92.98& 71.38& 52.71& 89.21 &91.74 &73.80& 58.99\\
&    & \online{} 
&{88.86} &{93.08}& \textbf{72.38} &{57.78}& \textbf{89.85}& \textbf{93.30}& \textbf{75.32} &\textbf{65.48}\\
&    & \batch{}
&\textbf{88.96}& \textbf{93.11}& {72.28}& \textbf{58.85} &{89.84} &\textbf{93.30} &{75.27} &{65.44}\\\hline
\multirow{6}{*}{\begin{sideways}{wsj} \end{sideways}} 
&\multirow{3}{*}{\begin{sideways}{l:small} \end{sideways}}
    & \noDA{}  
&94.64&95.44&83.38&82.72&95.73&95.88&\textbf{90.36}&87.87\\
&    & \online{}
&\textbf{94.86}&\textbf{95.53}&{85.37}&{85.22}&\textbf{95.80}&{96.21}&89.89&\textbf{89.70}\\
&    & \batch{} 
&{94.80}&{95.46}&\textbf{85.51}&\textbf{85.38}&\textbf{95.80}&\textbf{96.22}&89.89&\textbf{89.70}\\\cline{2-11}
&\multirow{3}{*}{\begin{sideways}{l:big} \end{sideways}}
    & \noDA{}
&96.44&\textbf{96.85}&92.75&85.38&96.56&96.72&93.35&88.04\\
&    & \online{} 
&\textbf{96.50}&\textbf{96.85}&\textbf{93.55}&{86.38}&{96.62}&\textbf{96.89}&{93.35}&{91.69}\\
&    & \batch{}
&{96.47}&96.82&{93.48}&\textbf{86.54}&\textbf{96.63}&\textbf{96.89}&\textbf{93.42}&\textbf{91.86}\\
\end{tabular}
}

\caption{\online{} / \batch{} accuracies are generally
  better
  than \noDA{} (see bold numbers) and improve with both more training data and
  more unlabeled data.}\tablabel{details}

\end{table*}

\online{}. Before tagging a test sentence, 
both left and right distributional vectors are updated via $\freq\left(\bigram(c_i, w)\right)$ += 1
for each appearance of bigram ``$c_i$ $w$'' in the sentence.
Then the sentence is tagged using the
updated word representations.  
As tagging progresses, the
distributional representations become increasingly specific to the
target domain (TD), converging to the representations that \batch{} uses at the end of the tagging process.

In all three modes,
suffix and shape features
are always  fully specified, for both
known and unknown words.

\section{Experimental results}\label{result}

\tabref{comp} 
compares performance on SANCL
for a number of baselines and
four versions of FLORS: S\&S,
\newcite{schnabel13flors}'s version of FLORS,
``S\&S (reimpl.)'', our reimplementation of that version, 
and \batch{} and \online{}, the two versions of FLORS we use
in this paper.
Comparing lines
``S\&S'' and 
``S\&S (reimpl.)'' in the table, we see that our
reimplementation of FLORS is comparable to S\&S's.
For the rest of this paper, our setup 
for
\batch{} and \online{}
differs from S\&S's in three
respects. (i) We use Gigaword as additional unlabeled
data. (ii) 
When we train a FLORS model, then 
the corpora that the word representations are derived from
do not include the test set.
The set of corpora used by S\&S for this purpose includes
the test set.
We make this change because application data may
not be available at training time in DA.
(iii) 
 The word representations used when the FLORS
model is trained are derived from all six SANCL
domains. This simplifies the experimental setup as we only
need to train a single model, not one per domain.
\tabref{comp} shows that  our setup with these three changes 
(lines \batch{} and \online{})
has state-of-the-art
performance on SANCL for domain adaptation (bold numbers).


\eat{

\footnote{For space
  reasons, we moved results for 
TDs answers and emails from Tables \ref{tab:comp} and
\ref{tab:details} to supplementary -- they show the same
patterns as  the other three out-of-domain TDs.}

}

\begin{table*}[!t]
\centering
{\footnotesize
\begin{tabular}{l@{\hspace{0.22cm}}l||l@{\hspace{0.22cm}}l|l@{\hspace{0.22cm}}l|l@{\hspace{0.22cm}}l|l@{\hspace{0.22cm}}l|l@{\hspace{0.22cm}}l|l@{\hspace{0.22cm}}l}
      &           & \multicolumn{4}{c|}{unknowns}&
  \multicolumn{4}{c|}{unseens}&
  \multicolumn{4}{c}{known words}\\
      &           &  \multicolumn{2}{c}{u:0} & \multicolumn{2}{c|}{u:big}&  \multicolumn{2}{c}{u:0} & \multicolumn{2}{c|}{u:big}&  \multicolumn{2}{c}{u:0} & \multicolumn{2}{c}{u:big}\\
      &           &\multicolumn{1}{c}{err}&\multicolumn{1}{c}{std}&\multicolumn{1}{c}{err}&\multicolumn{1}{c|}{std}&\multicolumn{1}{c}{err}&\multicolumn{1}{c}{std}&\multicolumn{1}{c}{err}&\multicolumn{1}{c|}{std}&\multicolumn{1}{c}{err}&\multicolumn{1}{c}{std}&\multicolumn{1}{c}{err}&\multicolumn{1}{c}{std}\\\hline\hline
\multirow{3}{*}{\begin{sideways}{l:small} \end{sideways}}
    & \noDA{}  
& $.3670^\dagger$ & .00085 & .2104 & .00081 & $.1659^\dagger$ & .00076 & .1084 & .00056 & $.1309^\dagger$ & .00056 & .0801 & .00042\\
      & \online{}    
& .3094 & .00160 & $.2102^*$ & .00093 & .1467 & .00120 & $.1086^*$ & .00074 & .1186 & .00095 & $.0802^*$ & .00048\\
     & \batch{}     
& $.3050^\dagger$ & .00143 & .2101 & .00083 & $.1646^\dagger$ & .00145 & .1076 & .00060 & $.1251^\dagger$ & .00103 & .0801 & .00040\\\hline
\multirow{3}{*}{\begin{sideways}{l:big} \end{sideways}}
& \noDA{} 
& $.1451^\dagger$ & .00114 & .1042 & .00100 & .0732 & .00052 & .0690 & .00042 & .0534 & .00027 & .0503 & .00025\\
     & \online{}    
& .1404 & .00125 & $.1037^*$ & .00098 & .0727 & .00051 & $.0689^*$ & .00051 & .0529 & .00031 & $.0502^*$ & .00031\\
     & \batch{}     
& $.1382^\dagger$ & .00140 & .1033 & .00112 & .0723 & .00065 & .0680 & .00062 & .0528 & .00033 & .0502 & .00031
\end{tabular}
}

\caption{Error rates (err) and standard deviations
  (std) for tagging.
$\dagger$ (resp.\ $*$): significantly different from
  \online{} error rate above\&below (resp.\ 
from ``u:0'' error rate to the left).
}\tablabel{results}
\end{table*}


\tabref{details} investigates the effect of sizes of labeled
and unlabeled data on performance of \online{} and \batch{}.
We report accuracy for all (ALL) tokens, for tokens
occurring in both
l:big and l:small (KN), tokens  
occurring in 
neither  l:big
nor l:small (OOV) and tokens ocurring in l:big, but not in
l:small (SHFT).\footnote{We cannot give the standard,
  single OOV evaluation number here since OOVs are different in
  different conditions, hence the breakdown into three measures.}
Except for some minor variations in a few
cases, both using more labeled data 
and using more unlabeled data
improves tagging accuracy for both \online{} and
\batch{}. \online{} and \batch{} are generally better or as
good as 
\noDA{} (in bold), always on ALL and OOV, and with a few exceptions
also on KN and SHFT.

\begin{figure}[!t]
\begin{tabular}{c}
\includegraphics[width=\panelfactor\textwidth]{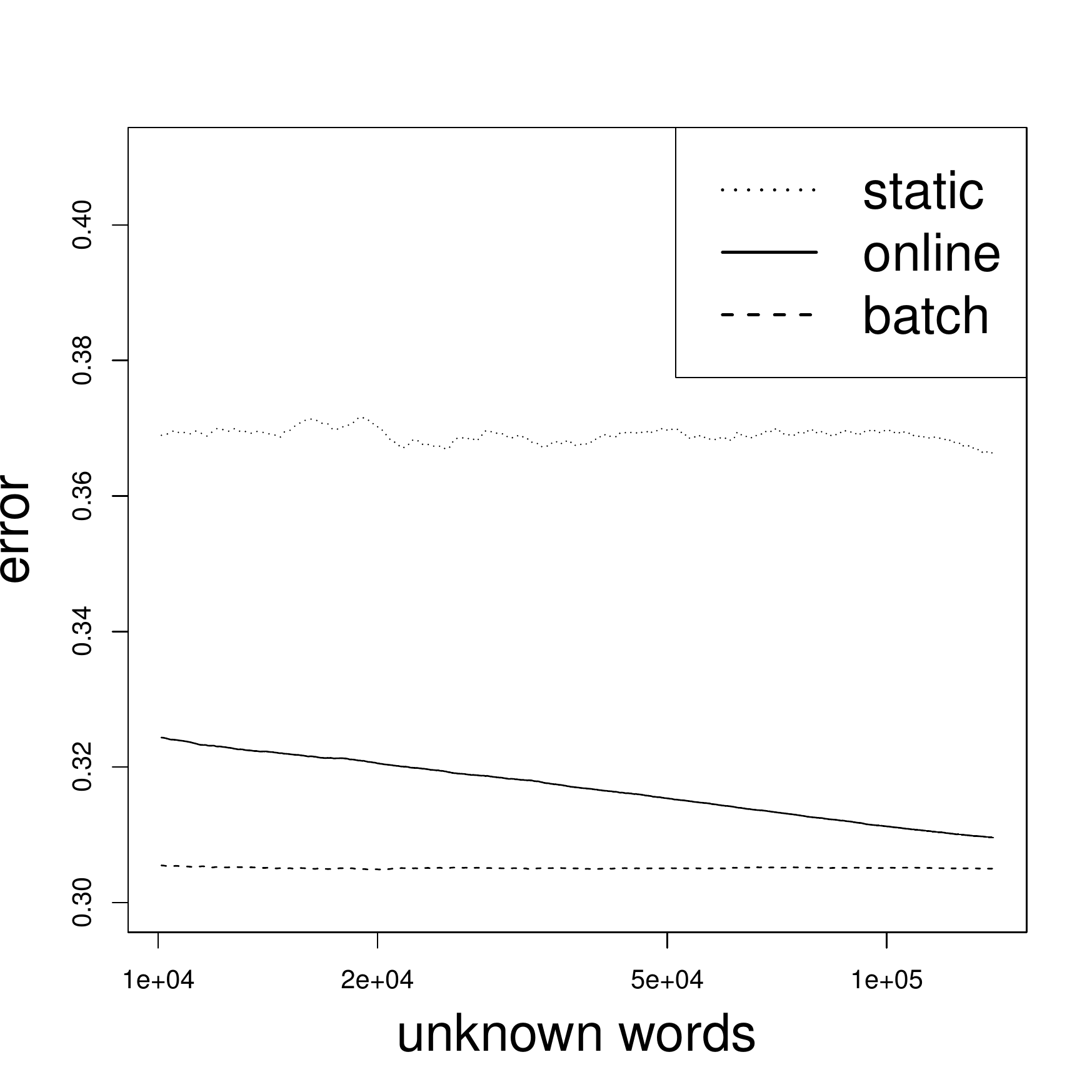}\\
\includegraphics[width=\panelfactor\textwidth]{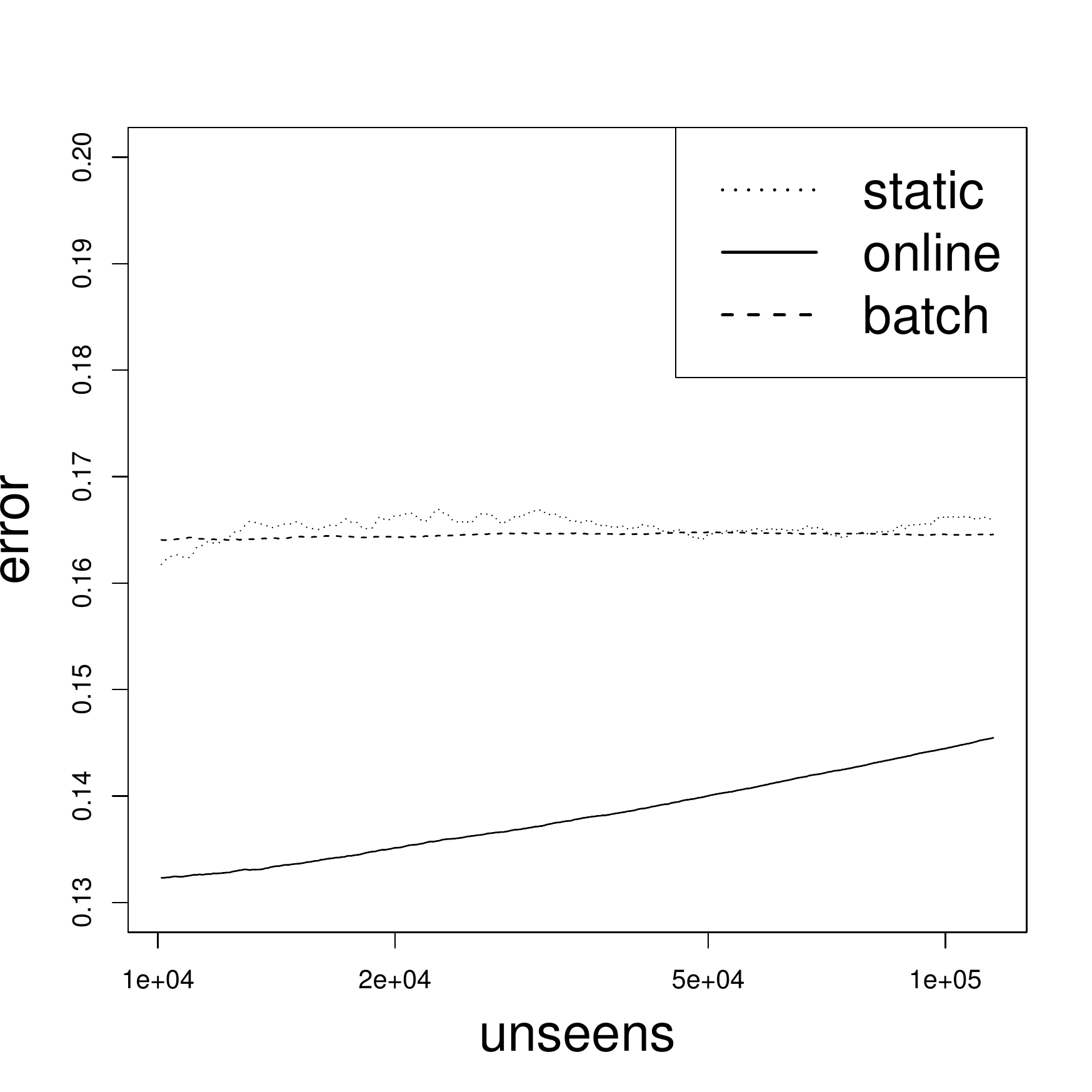}\\
\includegraphics[width=\panelfactor\textwidth]{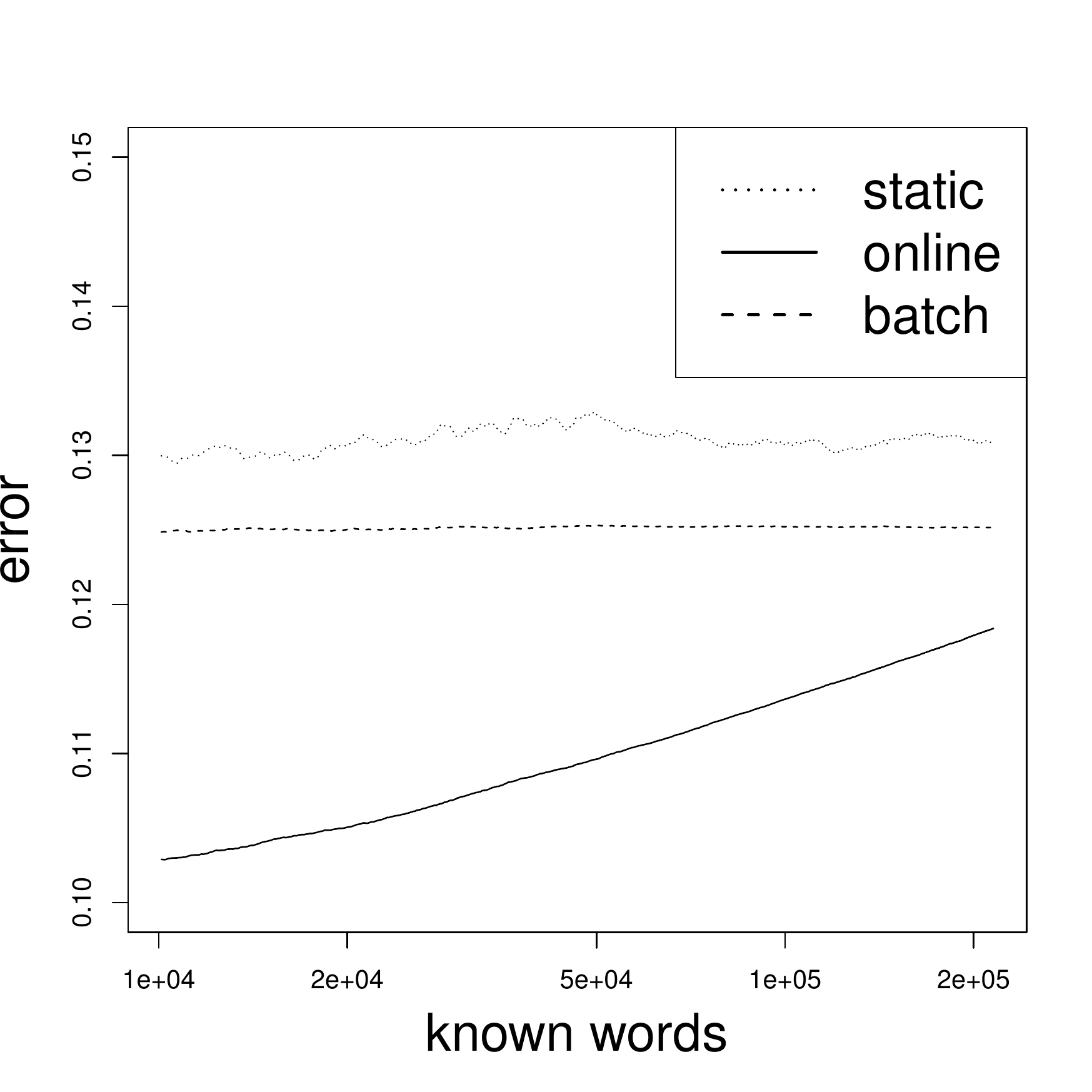}
\end{tabular}
  \caption{Error rates 
for unknown words, words with unseen tags and known words
for l:small/u:0. The x axis represents the number of
tokens of the respective type (e.g., number of tokens of
unknown words).}\figlabel{curve}
\end{figure}

\online{} performance is comparable to \batch{} performance:
it is slightly worse than \batch{} on u:0 (largest ALL
difference is .29) and at most .02 different from \batch{}
for ALL on u:big. We explain below why
\online{} is sometimes (slightly) \emph{better} than \batch{}, e.g.,
for ALL and condition 
l:small/u:big.


\enote{hs}{this point seems slightly less important to me
  since prior work has established that DA is better than
  \noDA{}. this paper is not about DA vs \noDA{}, but about \batch{}
  vs \online{} (but put this back in should we gain space)

As unlabeled data
increases (i.e., ``u:big''), ``\noDA{}'' only lags behind
slightly. This should be due to the fact that TD is much
smaller than source domain, and most words including unknown
words of TD actually can obtain good distributional
representations from u:big.
}

\subsection{Time course of tagging accuracy}

The \online{} model introduced in this paper has 
a property that is unique compared to most other work in
statistical NLP:
its \emph{predictions change} as it tags text because its
\emph{representations change.}


\enote{hs}{
A requirement of our study is the need to be able to
investigate the time course of tagging performance. This is
unique as machine learning models usually do
not change when unlabeled data is tagged. This is the
central contribution of this paper, so it is important to
investigate temporal effects (see \figref{curve}).  
}

To study this time course of changes, we need a large
application domain because subtle changes will be too
variable in the small test sets of the
SANCL TDs.
The only
labeled domain that is big enough is the
WSJ corpus.  We therefore reverse the standard setup and
train the model on the dev sets of the
five SANCL domains
(\emph{l:big}) or on the first 5000 labeled words of
reviews 
(\emph{l:small}). In this reversed setup, \emph{u:big} uses
the five unlabeled SANCL data sets and Gigaword as before.
Since
variance of performance
is  important, we
run on 100 randomly selected 50\%
samples of WSJ and report average and standard deviation of
tagging error over these 100 trials.


The results in \tabref{results}\footnote{Significance test:
  test of equal proportion, $p<.05$} show
that error rates are
only slightly worse for \online{}  than for \batch{}
 or the same. In fact, l:small/u:0 known  error
rate (.1186) is \emph{lower} for \online{} than for \batch{}
(similar to what we observed  in \tabref{details}).
This will be discussed at the
end of this section.

\tabref{results} includes results for ``unseens'' as well as 
unknowns because
\newcite{schnabel13flors} show that unseens
cause at least as many errors as unknowns. We define
\emph{unseens} as words with a tag that did not occur in training;
we compute unseen error
rates on \emph{all occurrences} of unseens, i.e.,
occurrences with both seen and unseen tags are included.
As \tabref{results} shows, the error rate for unknowns is
greater than for unseens which is in turn greater than the error rate on known words.

Examining the single conditions, we can see that \online{} fares better than \noDA{} 
in 10 out of 12 cases and only slightly worse for
l:small/u:big (unseens, known words: .1086 vs
.1084, .0802 vs .0801). In four conditions it is significantly
better with improvements ranging from .005 (.1404 vs .1451:
l:big/u:0, unknown words) to $>$.06 (.3094 vs .3670:
l:small/u:0, unknown words).

The differences between \online{} and \noDA{} in the other eight conditions  are
negligible. For the six u:big conditions, this is not surprising:
the Gigaword corpus
consists of news, so the large unlabeled data set is
in reality 
the same domain as WSJ. Thus,
if large unlabeled data sets are
available that are  similar to the TD, then one  might as well use \noDA{} tagging since the
extra work required for \online{}/\batch{} is unlikely to pay off.

Using more labeled data (comparing
l:small and l:big) always considerably decreases error
rates. We did not test for significance here because the
differences are so large. By the same token, 
using more unlabeled data (comparing u:0 and u:big) also
consistently decreases error rates. The differences are
large and significant for \online{} tagging in all six cases
(indicated by $*$ in the table).

There is no large
difference in variability \online{} vs.\ \batch{}
(see columns ``std''). Thus,
given that it has equal variability and higher performance,
\online{} is preferable to \batch{} since it assumes no
dataset
available prior to the start of tagging.

\figref{curve} shows the time course of tagging
  accuracy.\footnote{In response to a reviewer question,  the initial
    (leftmost) errors of \online{} and \noDA{} are \emph{not}
    identical; e.g., \online{} has a better chance of correctly tagging
    the very first occurrence of an unknown
    word because that very first occurrence has a meaningful (as opposed to random)
    distributed representation.}
\batch{} and \noDA{} have
constant error rates since they do not change representations during
tagging.
\online{} error decreases for unknown words -- approaching the
error rate of \batch{} -- as more and more is
learned with each additional occurrence of an unknown word
(top).

Interestingly, the error of \online{} \emph{increases} for
unseens and known words (middle\&bottom panels) (even
though it is always below the error rate of \batch{}).  The
reason is
that the \batch{} update swamps
the original training data for l:small/u:0 because the WSJ
test set is bigger by a large factor than the training
set. \online{} fares better here because in the beginning of
tagging the updates of the distributional representations
consist of small increments. We noticed this in
\tabref{details} too: there, \online{} outperformed
\batch{} in some cases on KN for 
l:small/u:big. In future work, we plan to investigate how to
weight distributional counts from the target data relative
to that from the (labeled und unlabeled) source data.

\section{Related work}\label{relatedwork}
Online learning
usually refers to \emph{supervised} learning algorithms that
update the model each time after processing a few training
examples. Many supervised learning algorithms 
are online 
or have online versions.
Active learning 
\cite{lewis94sequential,tong01active,laws11active}
is another supervised learning
framework that processes training examples -- 
usually obtained interactively -- in small batches
\cite{bordes05fast}.

\enote{hs}{cut to save space

by default,
can be run in
online mode or have online versions, including the perceptron
\cite{rosenblatt58perceptron}, backpropagation \cite{rhw86}
and MIRA \cite{crammer03ultraconservative} among many
others. 
}

All of this work on \emph{supervised online learning} is not
directly relevant to this paper since we address the problem
of \emph{unsupervised DA}. Unlike online
supervised learners, we keep the statistical model unchanged
during DA and adopt a representation learning
approach: each unlabeled context of a word is used to update
its representation.

\enote{hs}{cut to save space

The only example of unsupervised online 
DA we were able to find is
\cite{jain2011online}.
They postprocess the output of a face recognizer to ensure
that the same decision is made for two face candidates that
have similar confidence -- even if the decision boundary of
the original classifier separates them. 
The representation learning approach we adopt is potentially
more powerful than this type of postprocessing.

}

There is much work on 
unsupervised DA for POS tagging, including work using
constraint-based methods
\cite{subramanya10efficient,rush2012constraints}, instance
weighting \cite{choi2012selection}, self-training
\cite{huang2009self,huang10exploring}, and co-training
\cite{kubler2011fast}. All of this work uses batch
learning.
For space reasons, we do not discuss
supervised DA (e.g., 
\newcite{daume2006domain}).

\enote{hs}{cut unless we have space

\newcite{miller2006rapid}'s  POS tagging model uses
inflectional, derivational and orthographic features and
is not adapted during DA, similar to our work.
In fact, nothing
changes during DA, so the
approach is not really an instance of DA.

}

\section{Conclusion}\label{conclusion}
We introduced
online updating of word
representations, a new domain adaptation method 
for cases where target domain data are read from a
stream and \batch{} processing is not possible.
We showed that online unsupervised DA  performs as well as
batch learning. It also significantly lowers error
rates compared to \noDA{} (i.e., no domain adaptation). 
Our implementation of FLORS is
available at \url{cistern.cis.lmu.de/flors}


\textbf{Acknowledgments.}
This work was supported by 
a Baidu scholarship awarded to Wenpeng Yin and by
Deutsche Forschungsgemeinschaft (grant DFG SCHU 2246/10-1
FADeBaC).

\bibliographystyle{acl}
\bibliography{acl2015}

\enote{ts}{

- The abbreviation ``DA'' was never defined

- I don't know if starting with the definition of
unsupervised domain adaptation is the right way to get the
reader interested - maybe start with the example first.

- Another selling point of our method: It is very efficient,
as updates are super sparse. I think this should be
mentioned somewhere

- Maybe it's just me - but somehow the general take-home
message of the paper is neutral - slightly negative

- You included the URL for the Java implementation of
Liblinear - I would rather point to the original website:
http://www.csie.ntu.edu.tw/~cjlin/liblinear/

- Section 3: I would move the paragraph 
`` Let f(w) be the concatentation of the two distributional
and suffix and shape vectors of word w. Then FLORS
represents token v i as follows: f(v i−2 ) ⊕ f(v i−1 ) ⊕ f(v
i ) ⊕ f(v i+1 ) ⊕ f(v i+2 )  ..''
to appear before ``Distributional features.'' That way, the
discussion of why distributional features are stable across
domains also fits better.

- I would rewrite the sentence:
 Log weighting and normalization are applied to a copy of
 each distributional vector before it is used by LIBLINEAR
 in training or classification
as
Log weighting and normalization are applied to each count
vector before it is used by LIBLINEAR in training or
classification 

- I think one could be a little bit more explicit about what
the \batch{} scenario means. I think what you mean is you
trained the tagger once and then used this model for all
timesteps. Another interpretation would be to train a model
on all data up to time point I in a \batch{} fashion. 

- Is it possible to make Figure 1 span two columns? It looks
a little bit confusing that way

- I found this sentence a bit odd:
``We did some initial experiments with differential
weighting of source vs target counts, but leave this for
future work''
This raises more questions than it answers - maybe just
state that you leave it for future work

}

\end{document}